\documentclass[11pt]{article}
\usepackage{coling2018}
\usepackage{times}
\usepackage{url}
\usepackage{latexsym}
\usepackage{latexsym}
\usepackage{color}
\usepackage{multirow}
\usepackage{geometry}
\usepackage{pdflscape}

\title{Sentiment Analysis of Code-Mixed Indian Languages: \\ An Overview of SAIL\_Code-Mixed Shared Task @ICON-2017}

\author{Braja Gopal Patra, Dipankar Das, and Amitava Das\\
  University of Texas Health Science Center, Houston, Texas, USA \\
  Jadavpur University, Kolkata, India \\
  IIIT Sri City, Andhra Pradesh, India \\
  {\tt brajagopal.cse@gmail.com, dipankar.dipnil2005@gmail.com,}\\ {\tt amitava.das@iiits.in} \\}

\date{}

\begin{document}
\maketitle
\begin{abstract}
Sentiment analysis is essential in many real-world applications such as stance detection, review analysis, recommendation system, and so on. Sentiment analysis becomes more difficult when the data is noisy and collected from social media. India is a multilingual country; people use more than one languages to communicate within themselves. The switching in between the languages is called code-switching or code-mixing, depending upon the type of mixing. This paper presents overview of the shared task on sentiment analysis of code-mixed data pairs of Hindi-English and Bengali-English collected from the different social media platform. The paper describes the task, dataset, evaluation, baseline and participant's systems.
\end{abstract}

\section{Introduction}
\label{intro}
The past decade witnessed rapid growth and widespread usage of social media platforms by generating a significant amount of user-generated text. The user-generated texts contain high information content in the form of news, expression, or knowledge. Automatically mining information from user-generated data is unraveling a new field of research in Natural Language Processing (NLP) and has been a difficult task due to unstructured and noisy nature. In spite of the existing challenges, much research has been conducted on user-generated data in the field of information extraction, sentiment analysis, event extraction, user profiling and many more. 

According to Census of India, there are 22 scheduled languages and more than 100 non scheduled languages in India\footnote{http://www.censusindia.gov.in/Census\_Data\_2001/Census\_Data\_Online/Language/data\_on\_language.aspx}. There are 462 million internet users in India\footnote{https://www.statista.com/topics/2157/internet-usage-in-india/} and most people know more than one language. They express their feelings or emotions using more than one languages, thus generating a new code-mixed/code-switched language. The problem of code-mixing and code-switching are well studied in the field of NLP~\cite{sharma2014borrowing,das2014identifying}. Information extraction from Indian internet user-generated texts become more difficult due to this multilingual nature. Much research has been conducted in this field such as language identification~\cite{Banerjee:2014,mandal2015adaptive}, part-of-speech tagging~\cite{ghosh2016part}. Joshi et al.~\shortcite{JoshiPSV16} have performed sentiment analysis in Hindi-English (HI-EN) code-mixed data and almost no work exists on sentiment analysis of Bengali-English (BN-EN) code-mixed texts.

The Sentiment Analysis of Indian Language (Code-Mixed) (SAIL \_Code-Mixed)\footnote{http://www.dasdipankar.com/SAILCodeMixed.html} is a shared task at ICON-2017\footnote{https://ltrc.iiit.ac.in/icon2017/}. Two most popular code-mixed languages namely Hindi and Bengali mixed with English were considered for the sentiment identification task. A total of 40 participants registered for the shared task and only nine teams have submitted their predicted outputs. Out of nine unique submitted systems for evaluation, eight teams submitted fourteen runs for HI-EN dataset whereas seven teams submitted nine runs for BN-EN dataset. The training and test dataset were provided after annotating the languages and sentiment (positive, negative, and neutral) tags. The language tags were automatically annotated with the help of different dictionaries whereas the sentiment tags were manually annotated. The submitted systems are ranked using the macro average f-score.

The paper is organized as following manner. Section~\ref{related:work} describes the NLP in Indian languages mainly related to code-mixing and sentiment analysis. The detailed statistics of the dataset and evaluation are described in Section~\ref{data}. The baseline systems and participant's system description are described in Section~\ref{rns}. Finally, conclusion and future research are drawn in Section~\ref{conclusion}.

\section{Related Work}
\label{related:work}
With the rise of social media and user-generated data, information extraction from user-generated text became an important research area. Social media has become the voice of many people over decades and it has special relations with real time events. The multilingual user have tendency to mix two or more languages while expressing their opinion in social media, this phenomenon leads to generate a new code-mixed language. So far, many studies have been conducted on why the code-mixing phenomena occurs and can be found in Kim~\shortcite{kim2006reasons}.

Several experiments have been performed on social media texts including code-mixed data. The first step toward information gathering from these texts is to identify the languages present. Till date, several language identification experiments or tasks have been performed on several code-mixed language pairs such as Spanish-English~\cite{negron2009spanish,solorio2011analyzing}, French-English~\cite{voss2014finding}, Hindi-English~\cite{sharma2014borrowing,das2014identifying}, Hindi-English-Bengali~\cite{barman2014code}, Bengali-English~\cite{das2014identifying}. Many shared tasks have also been organized for language identification of code-mixed texts. \textit{Language Identification in Code-Switched Data}~\footnote{http://emnlp2014.org/workshops/CodeSwitch/call.html} was one of the shared tasks which covered four language pairs such as Spanish-English, Modern Standard Arabic and Arabic dialects, Chinese-English, and Nepalese-English. In the case of Indian languages, \textit{Mixed Script Information Retrieval}~\cite{sequiera2015overview} shared task at FIRE-2015\footnote{http://fire.irsi.res.in/fire/2015/home} was organized for eight code-mixed Indian languages such as Bangla, Gujarati, Hindi, Kannada, Malayalam, Marathi, Tamil, and Telugu mixed with English.

The second step is the identification of Part-of-Speech (POS) tags in code-mixed data and only handful of experiments have been performed in it such as Spanish-English~\cite{solorio2008part}, Hindi-English~\cite{vyas2014pos}. POS Tagging for Code-mixed Indian Social Media\footnote{http://amitavadas.com/Code-Mixing.html} shared task was organized for language pairs such as Bengali-English, Hindi-English, and Telugu-English. However, to best of the authors' knowledge no tasks on POS tagging were found on other code-mixed Indian languages. Again, Named Entity Recognition (NER) of code-mixed language\footnote{http://www.au-kbc.org/nlp/CMEE-FIRE2016/} shared task was organized for identifying named entities in Hindi-English and Tamil-English code-mixed data~\cite{rao2016cmee}.

Sentiment analysis or opinion mining from code-mixed data is one of the difficult tasks and the reasons are listed below. 
\begin{itemize}
\item Generally, code-mixed data is noisy in nature and it requires cleaning and normalization.
\item It needs several steps such as language identification and POS tagging.
\item There are no sentiment annotated code-mixed lexicon available for any language pairs.
\item The available code-mixed datasets are small in size to perform any unsupervised classification. 
\end{itemize}

Sentiment analysis of Hindi-English code-mixed was performed by Joshi et al.~\shortcite{JoshiPSV16} which used sub-word level representations in LSTM architecture to perform it. This is one of the initial tasks in sentiment analysis of HI-EN code-mixed dataset. There are several applications on code-mixed data which depends on sentiment analysis such as stance detection, aspect based sentiment analysis. However, there are several tasks available on sentiment analysis of Indian language tweets~\cite{patra2015shared,akhtar2016hybrid}. The shared task on sentiment analysis in Indian languages (SAIL) tweets\footnote{http://amitavadas.com/SAIL/data.html} focused on sentiment analysis of three Indian languages: Bengali, Hindi, and Tamil~\cite{patra2015shared}. 

\section{Dataset and Evaluation}
\label{data}
This section describes statistics of the dataset and the evaluation procedure. Preparing a gold standard dataset is the first step towards achieving good accuracy. Several tasks in the field of NLP suffer from lack of gold standard dataset. In the case of Indian languages, there is no such code-mixed dataset available for research purpose. Thus, we developed the dataset and the details are provided below.

\subsection{Dataset}
Data collection is a time consuming and tedious task in terms of human resource. Two code-mixed data pairs HI-EN and BN-EN are provided for developing sentiment analysis systems. The Twitter4j\footnote{http://twitter4j.org/en/} API was used to collect both Bengali and Hindi code-mixed data from Twitter. Initially, common Bengali and Hindi words were collected and then searched using the above API. The collected words are mostly sentiment words in Romanized format. Plenty of tweets had noisy words such as words from other languages and words in utf-8 format. 

After collection of code-mixed tweets, some were rejected. There are three reasons for which a tweet was rejected. 
\begin{itemize}
    \item a tweet is incomplete, i.e. there is not much information available in the tweet.
    \item a tweet is spam, advertisement or slang.
    \item a tweet does not have either Bengali or Hindi words.
\end{itemize}

The hashtags and urls are kept unchanged. Then words are automatically tagged with language information using a dictionary which is developed manually. Finally, tweets are manually annotated with the positive, negative, and neutral polarity. Missed language tags or wrongly annotated language tags are corrected manually during sentiment annotation.

Any of the six language tags is used to annotate the language to each of the words and these are HI (Hindi), EN (English), BN (Bengali), UN(Universal), MIX (Mix of two languages), EMT (emoticons). MIX words are basically the English words with Hindi or Bengali suffix, for example, \textit{Delhite} (\textit{in Delhi}). Sometimes, the words are joined together by mistake due to the typing errors, for example, \textit{jayegiTension} (\textit{tension will go away}). UN words are basically symbols, hashtags, or name etc. The statistics of training and test tweets for Bengali and Hindi code-mixed datasets are provided in Table~\ref{stat}. Some examples of HI-EN and BN-EN datasets with sentiment tags are given below. 

\begin{itemize}
    \item \textbf{BI-EN}: Irrfan Khan hollywood e abar dekha debe, trailer ta toh awesome ar acting o enjoyable. (\textbf{positive})\\
\textbf{Tagged}: Irrfan/EN Khan/EN hollywood/EN e/BN abar/BN dekha/BN debe/BN ,/UN trailer/EN ta/BN toh/BN awesome/EN ar/BN acting/EN o/BN enjoyable/EN ./UN \\
\textbf{Translation}: Irrfan Khan will be seen in Hollywood again, trailer is awesome and acting is also enjoyable.

\item \textbf{BI-EN}: Ei movie take bar bar dekheo er matha mundu kichui bojha jaye na. Everything boddo confusing and amar mote not up to the mark. (\textbf{negative})\\
\textbf{Tagged}: Ei/BN movie/EN take/BN bar/BN bar/BN dekheo/BN er/BN matha/BN mundu/BN kichui/BN bojha/BN jaye/BN na/BN ./UN Everything/EN boddo/BN confusing/EN and/EN amar/BN mote/BN not/EN up/EN to/EN the/EN mark/EN ./UN\\
\textbf{Translation}: After watching repeated times I can't understand anything. Everything is so confusing and I think its not up to the mark. 


\item \textbf{HI-EN}: bhai jan duaa hei k appki film sooper dooper hit ho (\textbf{positive}) \\
\textbf{Tagged}: bhai/HI jan/HI duaa/HI hei/HI k/HI appki/HI film/HI sooper/EN dooper/HI hit/HI ho/HI ./UN \\
\textbf{Translation}: Brother I pray that your film will be a super duper hit.

\item \textbf{HI-EN}: yaaaro yeah \#railbudget2015 kitne baaje start hooga ? (\textbf{neutral}) \\
\textbf{Tagged}: yaaaro/HI yeah/EN \#railbudget2015/EN kitne/HI baaje/HI start/EN hooga/EN ?/UN \\
\textbf{Translation}: Friends, when will \#railbudget2015 start?
\end{itemize}

\begin{table*}[!htbp]
\centering
\begin{tabular}{cccccc}
\hline
 & \textbf{Dataset} & \textbf{Positive} & \textbf{Negative} & \textbf{Neutral} & \textbf{Total} \\ \hline
\multirow{2}{*}{\textbf{Training}} & \textbf{HI-EN} & 4064 & 2972 & 5900 & 12936 \\ \cline{2-6} 
 & \textbf{BN-EN} & 1000 & 1000 & 500 & 2500 \\ \hline
\multirow{2}{*}{\textbf{Test}} & \textbf{HI-EN} & 1740 & 1247 & 2538 & 5525 \\ \cline{2-6} 
 & \textbf{BN-EN} & 1240 & 791 & 1007 & 3038 \\ \hline
\end{tabular}
\caption{\label{stat}Statistics of the training and test dataset with respect to sentiment}
\end{table*}


\subsection{Evaluation}
The precision, recall and f-score are calculated using the sklearn package of \textit{scikit-learn}~\cite{scikit-learn}. The macro average f-score is used to rank the submitted systems, because it independently calculates the metric for each classes and then takes the average hence treating all classes equally. 
Two different types of evaluation are considered and these are described below.

\begin{enumerate}
\item \textbf{Overall}: The macro average precision, recall, and f-score are calculated for all submitted runs. 
\item \textbf{Two way}: Then, two way classification approach is used where the system will be evaluated on two classes. For positive sentiment calculation, the predicted negative and neutral tags are converted to \textit{other} for both gold and predicted output by making the task as binary classification. Then, the macro averaged precision, recall, and f-score are calculated. Similar process is also applied for negative and neural metrics calculation.
\end{enumerate}

\section{System Description and Comparison}
\label{rns}

\subsection{Baseline}
The baseline systems are developed by randomly assigning any of the sentiment values to each of the test instances. Then, similar evaluation techniques are applied to the baseline system and the results are presented in Table~\ref{resultstable}.

\subsection{System Descriptions}
This subsection describes the details of systems submitted for the shared task. Six teams have submitted their system details and those are described below in order of decreasing f-score.

\begin{landscape}
\begin{table*}[!htbp]
\centering
\begin{tabular}{lccccccccccccc}
\cline{3-14}
\multicolumn{2}{l}{\multirow{2}{*}{}} & \multicolumn{12}{c}{\textbf{HI-EN}} \\ \cline{3-14} 
\multicolumn{2}{l}{} & \multicolumn{3}{c}{\textbf{Overall}} & \multicolumn{3}{c}{\textbf{Positive}} & \multicolumn{3}{c}{\textbf{Negative}} & \multicolumn{3}{c}{\textbf{Neutral}} \\ \hline
\multicolumn{1}{l}{\textbf{SystemID}} & \textbf{RunID} & \textbf{P} & \textbf{R} & \textbf{F} & \textbf{P} & \textbf{R} & \textbf{F} & \textbf{P} & \textbf{R} & \textbf{F} & \textbf{P} & \textbf{R} & \textbf{F} \\ \hline
\multicolumn{1}{l}{\textbf{IIIT-NBP}} & 2 & 0.597 & 0.560 & 0.569 & 0.719 & 0.699 & 0.707 & 0.691 & 0.644 & 0.659 & 0.674 & 0.671 & 0.663 \\ \hline
\multicolumn{1}{l}{\textbf{BIT Mesra}} & 1 & 0.573 & 0.558 & 0.564 & 0.701 & 0.697 & 0.699 & 0.679 & 0.655 & 0.664 & 0.651 & 0.652 & 0.650 \\ \hline
\multicolumn{1}{l}{\textbf{BIT Mesra}} & 2 & 0.573 & 0.559 & 0.564 & 0.701 & 0.697 & 0.699 & 0.679 & 0.657 & 0.666 & 0.652 & 0.653 & 0.651 \\ \hline
\multicolumn{1}{l}{\textbf{JU-KS}} & 1 & 0.579 & 0.556 & 0.562 & 0.691 & 0.696 & 0.693 & 0.695 & 0.646 & 0.662 & 0.659 & 0.660 & 0.657 \\ \hline
\multicolumn{1}{l}{\textbf{IIIT-NBP}} & 1 & 0.607 & 0.547 & 0.557 & 0.719 & 0.696 & 0.705 & 0.709 & 0.628 & 0.646 & 0.669 & 0.661 & 0.649 \\ \hline
\multicolumn{1}{l}{\textbf{NLP\_CEN\_AMRITA}} & 1 & 0.589 & 0.540 & 0.550 & 0.720 & 0.685 & 0.696 & 0.684 & 0.633 & 0.648 & 0.654 & 0.647 & 0.635 \\ \hline
\multicolumn{1}{l}{\textbf{CFILT}} & 2 & 0.591 & 0.516 & 0.524 & 0.730 & 0.669 & 0.682 & 0.674 & 0.608 & 0.622 & 0.656 & 0.637 & 0.615 \\ \hline
\multicolumn{1}{l}{\textbf{CFILT}} & 1 & 0.575 & 0.508 & 0.514 & 0.732 & 0.651 & 0.663 & 0.645 & 0.614 & 0.624 & 0.645 & 0.630 & 0.609 \\ \hline
\multicolumn{1}{l}{\textbf{Subway}} & 2 & 0.464 & 0.462 & 0.461 & 0.628 & 0.619 & 0.622 & 0.576 & 0.583 & 0.578 & 0.587 & 0.587 & 0.587 \\ \hline
\multicolumn{1}{l}{\textbf{Subway}} & 1 & 0.460 & 0.460 & 0.459 & 0.630 & 0.621 & 0.624 & 0.573 & 0.582 & 0.575 & 0.580 & 0.580 & 0.580 \\ \hline
\multicolumn{2}{l}{\textbf{Baseline}} & \textbf{0.337} & \textbf{0.338} & \textbf{0.331} & \textbf{0.343} & \textbf{0.500} & \textbf{0.407} & \textbf{0.387} & \textbf{0.500} & \textbf{0.436} & \textbf{0.270} & \textbf{0.500} & \textbf{0.351} \\ \hline
\multicolumn{1}{l}{\textbf{CEN@Amrita}} & 2 & 0.275 & 0.354 & 0.309 & 0.671 & 0.694 & 0.672 & 0.387 & 0.500 & 0.436 & 0.649 & 0.648 & 0.642 \\ \hline
\multicolumn{1}{l}{\textbf{CEN@Amrita}} & 1 & 0.430 & 0.339 & 0.300 & 0.650 & 0.665 & 0.653 & 0.721 & 0.503 & 0.444 & 0.622 & 0.617 & 0.605 \\ \hline
\multicolumn{1}{l}{\textbf{SVNIT}} & 1 & 0.105 & 0.333 & 0.160 & 0.157 & 0.500 & 0.240 & 0.387 & 0.500 & 0.436 & 0.270 & 0.500 & 0.351 \\ \hline
\multicolumn{1}{l}{\textbf{SVNIT}} & 2 & 0.105 & 0.333 & 0.160 & 0.157 & 0.500 & 0.240 & 0.387 & 0.500 & 0.436 & 0.270 & 0.500 & 0.351 \\ \hline
\multicolumn{2}{l}{} & \multicolumn{12}{c}{\textbf{BN-EN}} \\ \hline
\multicolumn{1}{l}{\textbf{IIIT-NBP}} & 2 & 0.551 & 0.534 & 0.526 & 0.664 & 0.633 & 0.633 & 0.683 & 0.666 & 0.673 & 0.633 & 0.65 & 0.621 \\ \hline
\multicolumn{1}{l}{\textbf{IIIT-NBP}} & 1 & 0.552 & 0.531 & 0.524 & 0.670 & 0.640 & 0.641 & 0.678 & 0.655 & 0.664 & 0.632 & 0.649 & 0.619 \\ \hline
\multicolumn{1}{l}{\textbf{NLP\_CEN\_AMRITA}} & 1 & 0.517 & 0.516 & 0.513 & 0.641 & 0.631 & 0.633 & 0.654 & 0.655 & 0.654 & 0.613 & 0.622 & 0.615 \\ \hline
\multicolumn{1}{l}{\textbf{JU\_KS}} & 1 & 0.606 & 0.524 & 0.504 & 0.695 & 0.623 & 0.613 & 0.737 & 0.658 & 0.677 & 0.643 & 0.644 & 0.573 \\ \hline
\multicolumn{1}{l}{\textbf{CFILT}} & 1 & 0.528 & 0.476 & 0.455 & 0.638 & 0.597 & 0.588 & 0.667 & 0.606 & 0.616 & 0.613 & 0.618 & 0.557 \\ \hline
\multicolumn{1}{l}{\textbf{CFILT}} & 2 & 0.538 & 0.478 & 0.447 & 0.659 & 0.587 & 0.565 & 0.662 & 0.619 & 0.630 & 0.617 & 0.616 & 0.545 \\ \hline
\multicolumn{2}{l}{\textbf{Baseline}} & \textbf{0.342} & \textbf{0.343} & \textbf{0.339} & \textbf{0.296} & \textbf{0.500} & \textbf{0.372} & \textbf{0.370} & \textbf{0.500} & \textbf{0.425} & \textbf{0.334} & \textbf{0.500} & \textbf{0.401} \\ \hline
\multicolumn{1}{l}{\textbf{AMRITA\_CEN}} & 1 & 0.322 & 0.340 & 0.318 & 0.411 & 0.438 & 0.408 & 0.470 & 0.462 & 0.451 & 0.608 & 0.611 & 0.609 \\ \hline
\multicolumn{1}{l}{\textbf{CEN@Amrita}} & 1 & 0.230 & 0.309 & 0.258 & 0.570 & 0.569 & 0.569 & 0.370 & 0.500 & 0.425 & 0.616 & 0.627 & 0.579 \\ \hline
\multicolumn{1}{l}{\textbf{SVNIT}} & 1 & 0.136 & 0.333 & 0.193 & 0.204 & 0.500 & 0.290 & 0.370 & 0.500 & 0.425 & 0.334 & 0.500 & 0.401 \\ \hline
\end{tabular}
\caption{\label{resultstable}SAIL\_CodeMixed 2017 rankings ordered by F-Score of Overall Systems. (P: Precision, R: Recall, F: F-score)}
\end{table*}

\end{landscape}

\textbf{IIIT-NBP} team used features like GloVe word embeddings with 300 dimension and TF-IDF scores of word n-grams (one-gram, two-grams and tri-grams) as well as character n-grams (n varying from 2 to 6). Sklearn~\cite{scikit-learn} package is used to calculate the TF-IDF. Finally, two classifiers: ensemble voting (consisting of three classifiers - linear SVM, logistic regression and random forests) and linear SVM are used for classification. 

\textbf{JU\_KS} team used n-gram and sentiment lexicon based features. Small sentiment lexicons are manually prepared for both English and Bengali words. However, no sentiment lexicon is used for Hindi language. Bengali sentiment lexicon consists of a collection of 1700 positive and 3750 negative words whereas English sentiment lexicon consists of 2006 positive and 4783 negative words. Finally, Na\"ive Bayes multinomial is used to classify and system results are presented in Table~\ref{resultstable}.

\textbf{BIT Mesra} team submitted systems for only HI-EN dataset. During preprocessing, they removed words having UN language tags, URLs, hashtags and user mentions. An Emoji dictionary was prepared with sentiment tags. Finally, they used SVM and Na\"ive Bayes classifiers on uni-gram and bi-gram features to classify sentiment of the code-mixed HI-EN dataset only. 

\textbf{NLP\_CEN\_AMRITA} team have used different distributional and distributed representation. They used Document Term Matrix with N-gram varying from 1 to 5 for the representation and Support Vector Machines (SVM) as a classifier to make the final prediction. Their system performed well for n-grams range 5 and minimum document frequency 2 using the linear kernel. 

\textbf{CFIL} team uses simple deep learning for sentiment analysis on code-mixed data. The fastText\footnote{https://research.fb.com/fasttext/} tool is used to create word embeddings on sentiment corpus. Additionally, Convolutional Neural Networks was used to extract sub-word features. Bi-LSTM layer is used on word embedding and sub-word features together with max-pooling at the output which is again sent to a softmax layer for prediction. No additional features are used and hyper-parameters are selected after dividing training corpus to 70\% and 30\%.

\textbf{Subway} team submitted systems for HI-EN dataset only. Initially, words other than HI and EN tags are removed during the cleaning process. Then, a dictionary with bi-grams and tri-grams are collected from training data and sentiment polarity is annotated manually. TF-IDF scores for each matched n-grams are calculated and weights of 1.3 and 0.7 are assigned to bi-grams and tri-grams, respectively. Finally, Na\"ive Bayes classifier is used to get the sentiment. 

\subsection{Results and Discussion}
The baseline systems achieved better scores compared to \textit{CEN@AMRIT} and \textit{SVNIT} teams for HI-EN dataset; and \textit{AMRITA\_CEN}, \textit{CEN@Amrita} and \textit{SVNIT} teams for BN-EN dataset. \textit{IIIT-NBP} team has achieved the maximum macro average f-score of 0.569 across all the sentiment classes for HI-EN dataset. \textit{IIIT-NBP} also achieved the maximum macro average f-score of 0.526 for BN-EN dataset. Two way classification of HI-EN dataset achieved the maximum macro average f-score of 0.707, 0.666, and 0.663 for positive, negative, and neutral, respectively. Similarly, the two way classification of BN-EN dataset achieved the maximum average f-score of 0.641, 0.677, and 0.621 for positive, negative, and neutral, respectively. Again, the f-measure achieved using HI-EN dataset is better than BN-EN. The obvious reason for such result is that there are more instances in HI-EN than BN-EN dataset.

Most of the teams used the n-gram based features and it resulted in better macro average f-score. Most teams used the \textit{sklearn} for identifying n-grams. \textit{IIITH-NBP} team is only team to use character n-grams. Word embeddings is another important feature used by several teams. For word embeddings, Gensim\footnote{https://radimrehurek.com/gensim/} and fastText are used. \textit{JU\_KS} team has used sentiment lexicon based features for BN-EN dataset only. \textit{BITMesra} team has used emoji dictionary annotated with sentiment. Hashtags are considered to be one of the most important features for sentiment analysis~\cite{patramultilevel}, however they removed hashtags during sentiment identification.

Apart from the features, most of the teams used machine learning algorithms like SVM, Na\"ive Bayes. It is observed that the deep learning models are quite successful for many NLP tasks. \textit{CFIL} team have used the deep learning framework however the deep learning based system did not perform well as compared to machine learning based system. The main reason for the above may be that the training datasets provided are not sufficient to built a deep learning model.

\section{Conclusion and Future Work}
\label{conclusion}
This paper presents the details of shared task held during the ICON 2017. The competition presents the sentiment identification task from HI-EN and BN-EN code-mixed datasets. A random baseline system obtained macro average f-score of 0.331 and 0.339 for HI-EN and BN-EN datasets, respectively. The best performing team obtained maximum macro average f-score of 0.569 and 0.526 for HI-EN and BN-EN datasets, respectively. The team used word and character level n-grams as features and SVM for sentiment classification. We plan to enhance the current dataset and include more data pairs in the next version of the shared task. In future, more advanced task like aspect based sentiment analysis and stance detection can be performed on code-mixed dataset.


\begin{thebibliography}{}

\bibitem[\protect\citename{Akhtar \bgroup et al. \egroup }2016]{akhtar2016hybrid}
 Md Shad Akhtar, Ayush Kumar, Asif Ekbal and Pushpak Bhattacharyya.
\newblock 2016
\newblock A hybrid deep learning architecture for sentiment analysis.
\newblock {\em Proceedings of the 26th International Conference on Computational Linguistics: Technical Papers}, 
  pp. 482--493.

\bibitem[\protect\citename{Bali \bgroup et al.\egroup }2014]{sharma2014borrowing}
 Kalika Bali, Jatin Sharma, Monojit Choudhury and Yogarshi Vyas. 
\newblock 2014.
\newblock ``I am borrowing ya mixing?'' An Analysis of English-Hindi Code Mixing in Facebook.
\newblock {\em Proceedings of the First Workshop on Computational Approaches to Code Switching}, 
 pp. 116--126. 

 \bibitem[\protect\citename{Banerjee \bgroup et al.\egroup }2014]{Banerjee:2014}
 Somnath Banerjee, Alapan Kuila, Aniruddha Roy Sudip~K. Naskar, Paolo Rosso and Sivaji Bandyopadhyay.
\newblock 2014.
\newblock A Hybrid Approach for Transliterated Word-Level Language Identification: CRF with Post-Processing Heuristics.
\newblock {\em Proceedings of the Forum for Information Retrieval Evaluation}, 
pp. 54--59.   


\bibitem[\protect\citename{Barman \bgroup et al.\egroup }2014]{barman2014code}
 Utsab Barman, Amitava Das, Joachim Wagner and Jennifer Foster. 
\newblock 2014.
\newblock Code mixing: A challenge for language identification in the language of social media.
\newblock {\em Proceedings of the First Workshop on Computational Approaches to Code Switching}, 
 pp. 13--23. 

\bibitem[\protect\citename{Das and Gamb\"ack }2014]{das2014identifying}
 Amitava Das and Bj{\"o}rn Gamb{\"a}ck.
\newblock 2014.
\newblock Identifying languages at the word level in code-mixed Indian social media text.
\newblock {\em Proceedings of the 11th International Conference on Natural Language Processing}, 
 pp. 378--387.

 \bibitem[\protect\citename{Ghosh \bgroup et al.\egroup }2016]{ghosh2016part}
Souvick Ghosh, Satanu Ghosh and Dipankar Das.
\newblock 2016.
\newblock Part-of-speech Tagging of Code-Mixed Social Media Text.
\newblock {\em Proceedings of the Second Workshop on Computational Approaches to Code Switching},
  pp. 90--97.

  \bibitem[\protect\citename{Goldbarg }2009]{negron2009spanish}
Rosalyn~N. Goldbarg.
\newblock 2009.
\newblock Spanish-English codeswitching in email communication.
\newblock {\em Language@ internet},
  6(3):1--21.

\bibitem[\protect\citename{Joshi \bgroup et al.\egroup }2016]{JoshiPSV16}
Aditya Joshi, Ameya Prabhu, Manish Shrivastava and Vasudeva Varma.
\newblock 2016.
\newblock Towards Sub-Word Level Compositions for Sentiment Analysis of Hindi-English Code Mixed Text.
\newblock {\em Proceedings of the 26th International Conference on Computational Linguistics (COLING 2016)},
  pp. 2482--2491.

\bibitem[\protect\citename{Kim }2006]{kim2006reasons}
Eunhee Kim.
\newblock 2006.
\newblock Reasons and motivations for code-mixing and code switching.
\newblock {\em Issues in EFL},
  4(1):43--61.

\bibitem[\protect\citename{Mandal \bgroup et al.\egroup }2015]{mandal2015adaptive}
 Soumik Mandal,  Somnath Banerjee, Sudip~K. Naskar, Paolo Rosso and Sivaji Bandyopadhyay
\newblock 2015.
\newblock Adaptive Voting in Multiple Classifier Systems for Word Level Language Identification.
\newblock {\em Proceedings of the Forum for Information Retrieval Evaluation}.


\bibitem[\protect\citename{Patra \bgroup et al.\egroup }2015]{patra2015shared}
Braja~G. Patra, Dipankar Das, Amitava Das, and Rajendra Prasath. 
\newblock 2015.
\newblock Shared task on sentiment analysis in indian languages (SAIL) tweets-an overview.
\newblock {\em Proceedings of the International Conference on Mining Intelligence and Knowledge Exploration},
  pp. 650--655, Springer.
  
  
\bibitem[\protect\citename{Patra \bgroup et al.\egroup }2016]{patramultilevel}
Braja~G. Patra, Soumadeep Mazumdar, Dipankar Das, Paolo Rosso, and Sivaji Bandyopadhyay. 
\newblock 2016.
\newblock A Multilevel Approach to Sentiment Analysis of Figurative Language in Twitter.
\newblock {\em Proceedings of the 17th International Conference on Intelligent Text Processing and Computational Linguistics},
 Springer.

 \bibitem[\protect\citename{Pedregosa \bgroup et al. \egroup}2011]{scikit-learn}
Fabian Pedregosa, Ga\"el Varoquaux, Alexandre Gramfort, Vincent Michel, Bertrand Thirion, Olivier Grisel, Mathieu Blondel, Peter Prettenhofer, Ron Weiss, Vincent Dubourg, Jake Vanderplas, Alexandre Passos, David Cournapeau, Matthieu Brucher, Matthieu Perrot, \'Edouard Duchesnay.
\newblock Scikit-learn: Machine Learning in {P}ython
\newblock 2011.
\newblock {\em Journal of Machine Learning Research}, 
    12:2825--2830.
  
  
  
 


\bibitem[\protect\citename{Rao and Lalitha Devi }2016]{rao2016cmee}
Pattabhi RK Rao and Sobha Lalitha Devi
\newblock 2016.
\newblock CMEE-IL: Code Mix Entity Extraction in Indian Languages from Social Media Text@ FIRE 2016-An Overview. 
\newblock {\em Proceedings of the Forum for Information Retrieval Evaluation}, 
pp. 289--295.
  
  
\bibitem[\protect\citename{Royal \bgroup et al.\egroup }2015]{sequiera2015overview}
 Royal Sequiera, Monojit Choudhury, Parth Gupta, Paolo Rosso, Shubham Kumar, Somnath Banerjee, Sudip~K. Naskar and Bandyopadhyay, Sivaji Bandyopadhyay, Gokul Chittaranjan, Amitava Das, Kunal Chakma.
\newblock 2015.
\newblock Overview of FIRE-2015 Shared Task on Mixed Script Information Retrieval.
\newblock {\em Proceedings of the Forum for Information Retrieval Evaluation}, 
 pp. 19--25. 






\bibitem[\protect\citename{Solorio \bgroup et al.\egroup }2011]{solorio2011analyzing}
Thamar Solorio, Melissa Sherman, Yang Liu, Lisa~M. Bedore, Elisabeth~D. Pe{\~n}a and Aquiles Iglesias.
\newblock 2011.
\newblock Analyzing language samples of Spanish--English bilingual children for the automated prediction of language dominance.
\newblock {\em Natural Language Engineering},
  17(3):367--395, Cambridge University Press.
  
\bibitem[\protect\citename{Thamar and Yang}2008]{solorio2008part}
Thamar Solorio and Yang Liu. 
\newblock 2008.
\newblock Part-of-speech tagging for English-Spanish code-switched text.
\newblock {\em Proceedings of the Conference on Empirical Methods in Natural Language Processing},
  pp. 1051--1060, Association for Computational Linguistics.

 
 \bibitem[\protect\citename{Vyas \bgroup et al.\egroup }2014]{vyas2014pos}
 Yogarshi Vyas, Spandana Gella, Jatin Sharma, Kalika Bali and Monojit Choudhury.
\newblock 2014.
\newblock POS Tagging of English-Hindi Code-Mixed Social Media Content.
\newblock {\em Proceedings of the Empirical Methods in Natural Language Processing}, 
 pp. 974--979. 

 \bibitem[\protect\citename{Voss \bgroup et al.\egroup }2014]{voss2014finding}
 Clare~R. Voss, Stephen Tratz, Jamal Laoudi and Douglas~M. Briesch. 
\newblock 2014.
\newblock Finding Romanized Arabic Dialect in Code-Mixed Tweets.
\newblock {\em Proceedings of the Language Resources and Evaluation Conference}, 
 pp. 2249--2253. 
 

\end{thebibliography}

\end{document}